\documentclass[lettersize,journal]{IEEEtran}
\usepackage{amsmath,amsfonts}
\usepackage{algorithm}
\usepackage{algpseudocode}
\usepackage{array}
\usepackage{textcomp}
\usepackage{stfloats}
\usepackage{url}
\usepackage{verbatim}
\usepackage{graphicx}
\usepackage{cite}

\usepackage{hyperref}
\hypersetup{
    colorlinks=true,
    linkcolor=blue,
    urlcolor=blue
}

\usepackage{booktabs}
\usepackage{multirow}
\usepackage{pifont}
\usepackage{subcaption}
\usepackage{bm}

\begin{document}

\title{MGCA-Net: Multi-Grained Category-Aware Network for Open-Vocabulary Temporal Action Localization}

\author{Zhenying Fang, and Richang Hong,~\IEEEmembership{Senior Member,~IEEE}
\thanks{This work was supported by National Natural Science Foundation of China under Grant U23B2031. (Corresponding author: Richang Hong.)}
\thanks{Z. Fang and R. Hong are with the School of Computer Science and Information Engineering, Hefei University of Technology, 188065, China. e-mail: {zhenyingfang@mail.hfut.edu.cn, hongrc@hfut.edu.cn}.}
\thanks{Manuscript received April 19, 2021; revised August 16, 2021.}
}

\markboth{Journal of \LaTeX\ Class Files,~Vol.~14, No.~8, August~2021}%
{Shell \MakeLowercase{\textit{et al.}}: A Sample Article Using IEEEtran.cls for IEEE Journals}

\IEEEpubid{\parbox{\textwidth}{Copyright~\copyright~2026 IEEE. Personal use of this material is permitted.However, permission to use this material for any other purposes must be obtained from the IEEE by sending an email to pubs-permissions@ieee.org.}}

\maketitle

\begin{abstract}
Open-Vocabulary Temporal Action Localization (OV-TAL) aims to recognize and localize instances of any desired action categories in videos without explicitly curating training data for all categories. Existing methods mostly recognize action categories at a single granularity, which degrades the recognition accuracy of both base and novel action categories. To address these issues, we propose a Multi-Grained Category-Aware Network (MGCA-Net) comprising a localizer, an action presence predictor, a conventional classifier, and a coarse-to-fine classifier.
Specifically, the localizer localizes category-agnostic action proposals. For these action proposals, the action presence predictor estimates the probability that they belong to an action instance. At the same time, the conventional classifier predicts the probability of each action proposal over base action categories at the snippet granularity. Novel action categories are recognized by the coarse-to-fine classifier, which first identifies action presence at the video granularity—i.e., all action categories occurring in each input video—yielding coarse categories. Finally, it assigns each action proposal to one category from the coarse categories at the proposal granularity. Through coarse-to-fine category awareness for novel actions and the conventional classifier's awareness of base actions, multi-grained category awareness is achieved, effectively enhancing localization performance. Comprehensive evaluations on the THUMOS'14 and ActivityNet-1.3 benchmarks demonstrate that our method achieves state-of-the-art performance. Furthermore, our MGCA-Net achieves state-of-the-art results under the Zero-Shot Temporal Action Localization (ZS-TAL) setting. Our code is available at \url{https://github.com/zhenyingfang/MGCA-Net}.
\end{abstract}


\section{Introduction}\label{sec:introduction}

\begin{figure}[h]
  \centering
  \includegraphics[width=1.0\linewidth]{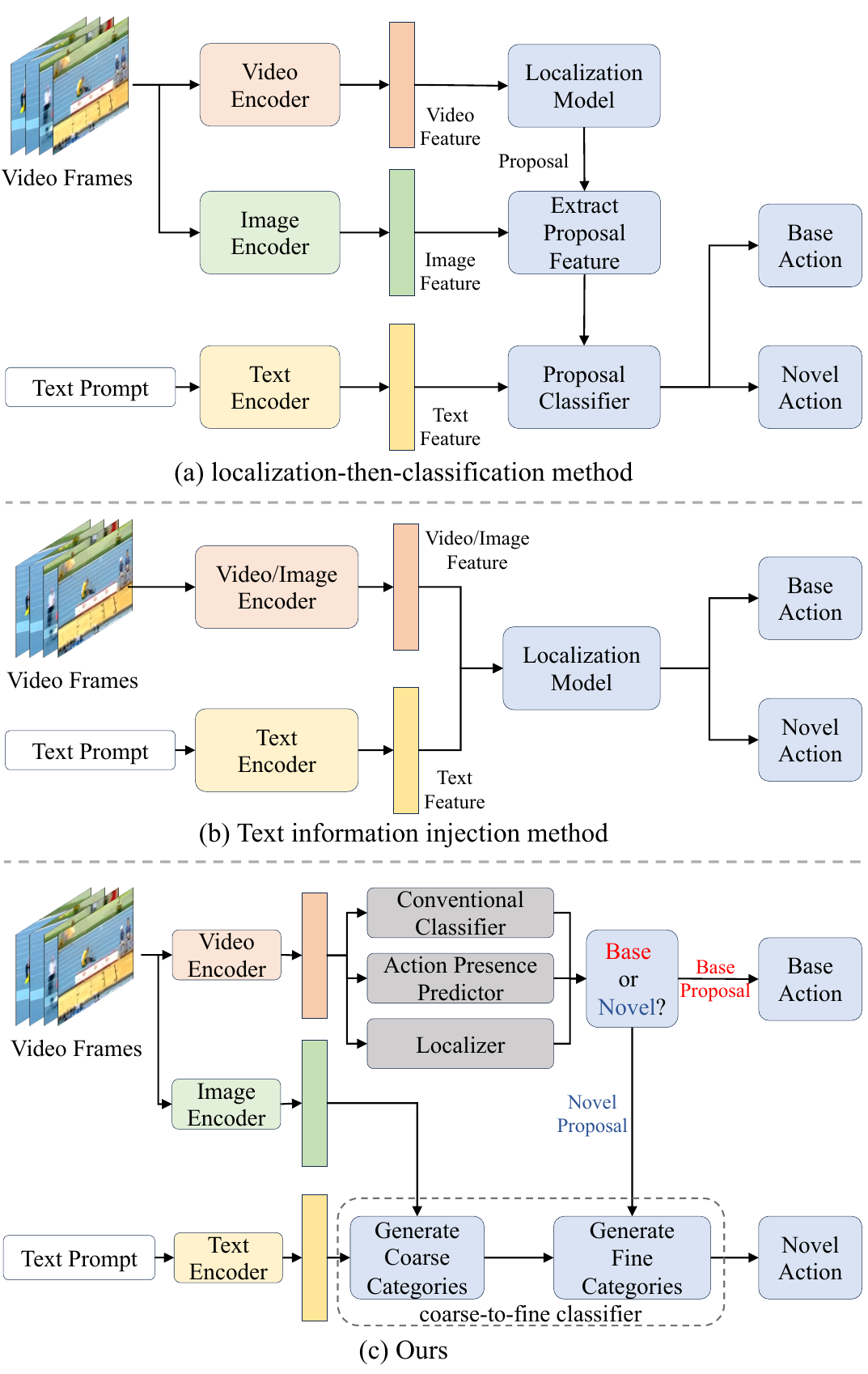}
  \caption{The structural comparison of (a) the localization-then-classification method, (b) the text information injection method, and (c) our proposed method.}
  \label{fig:method_cmp}
\end{figure}

\IEEEPARstart{T}{emporal} Action Localization (TAL) is a fundamental task in video understanding, whose goal is to predict the action category and temporal boundaries of each action instance in an untrimmed video. Traditional TAL methods~\cite{pcpcad2019, chen2021tcsvt, actionformer2022, tridet2023, cao2023tcsvt, lpr2023MMSJ, bdrcnet2025, brtal2025tcsvt, bfstal2025tcsvt, semtad2025tcsvt} follow a supervised learning paradigm and assume that the action categories within the training and testing sets remain identical. Nonetheless, this assumption confines the applicability of TAL to new and diverse scenarios, often necessitating model re-training to accommodate novel actions. Thus, developing TAL models capable of generalizing to any target actions beyond preset ones has been a long-standing goal.

To achieve this goal, researchers have explored multiple directions. Open-set TAL~\cite{opental2022CVPR} aims to localize all actions by assigning base categories and labeling novel actions as "unknown". Open-world TAL~\cite{owtal2023pr} extends open-set TAL by integrating continuous learning, allowing the model to be updated using annotations of novel actions after the initial training phase. However, these settings are not suitable for detecting novel actions.

\IEEEpubidadjcol

To detect novel actions, Open-Vocabulary TAL (OV-TAL) has been proposed, aiming to localize both the base action categories defined during the training phase and the novel action categories during inference. Existing OV-TAL methods can be categorized into two types: localization-then-classification and text information injection. As shown in Fig.~\ref{fig:method_cmp}. (a), the localization-then-classification methods~\cite{effprompt2022eccv, detal2024tpami, zeetad2024wacv, stovtal2025wacv}, which perceive action categories at the proposal granularity, first localize category-agnostic proposals, then extract proposal-level features and perform action classification leveraging the zero-shot capability of vision-language models (VLMs). Text information injection methods~\cite{stable2022eccv, unloc2023iccv, mprotea2024tcsvt, tifad2024neurips, ovformer2024bmvc}, which perceive action categories at the snippet granularity, inject textual features extracted by VLMs into the localization model to perform classification and localization simultaneously, as illustrated in Fig.~\ref{fig:method_cmp}. (b). Although both methods have achieved promising performance, they share a key limitation: they both perceive action categories at a single granularity, which restricts their capability to perceive them. To address this issue, we propose a Multi-Grained Category-Aware Network (MGCA-Net), as shown in Fig.~\ref{fig:method_cmp}. (c), which comprises a localizer, an action presence predictor, a conventional classifier, and a coarse-to-fine classifier.

1) The localizer is a common component in TAL for generating category-agnostic action proposals, which predicts the start and end times of the corresponding action proposal for each video snippet, where each video snippet typically consists of 16 consecutive frames from the input video~\cite{actionformer2022, tridet2023, bdrcnet2025}.

2) The conventional classifier and action presence predictor classify category-agnostic action proposals into base action instances (belonging to base actions) and novel proposals (belonging to novel actions). Specifically, the action presence predictor is trained in a category-agnostic manner to predict an action presence score (APS), which assesses the probability that a proposal represents an action instance. The conventional classifier is trained via traditional supervised learning using annotations from the training set, yielding classification probabilities for each proposal across base action categories. We retain all action proposals with an APS exceeding a threshold. Among the remaining proposals, if the maximum predicted probability of the conventional classifier for a proposal exceeds the threshold, its category is determined by the base action corresponding to this maximum probability, resulting in a base action instance and thereby achieving category awareness at the snippet granularity. Otherwise, novel proposals, whose categories are determined by the coarse-to-fine classifier, are obtained.

3) The coarse-to-fine classifier leverages the zero-shot capabilities of VLMs and employs a hierarchical process to recognize novel action categories. Specifically, it first predicts all possible action categories in the input video (i.e., coarse categories) by aligning text features and image features extracted by VLMs, thereby achieving category awareness at the video granularity. Subsequently, proposal-level features are derived from both the novel proposals and the image features. Finally, through contrastive learning, each proposal is assigned an action category by matching its proposal-level features against the previously predicted coarse categories, achieving category awareness at the proposal granularity.

By leveraging the coarse-to-fine classifier for recognizing novel action categories and the conventional classifier for recognizing base action categories, MGCA-Net achieves effective multi-grained category awareness. This capability improves the recognition accuracy of both base and novel actions. Further, it enhances localization performance, enabling MGCA-Net to achieve state-of-the-art performance in both OV-TAL and ZS-TAL tasks. Our contributions are summarized as:

\begin{itemize}
    \item We propose the Multi-Grained Category-Aware Network (MGCA-Net) to alleviate the low category recognition accuracy caused by single-granularity category awareness.
    \item We propose a conventional classifier to predict base action categories at the snippet granularity, and further propose an action presence predictor to divide action proposals into base or novel proposals.
    \item We propose a coarse-to-fine classifier to progressively identify novel action categories at the video and proposal granularities, thereby enabling coarse-to-fine recognition of novel actions.
    \item Extensive experiments across multiple benchmarks demonstrate that our method achieves state-of-the-art performance in OV-TAL and ZS-TAL settings.
\end{itemize}

\section{Related Works}\label{sec:related_works}

In this section, we first review previous work on the vision-language models and temporal action localization. Then, we review the methods for the OV-TAL task.

\subsection{Vision-Language Models}\label{subsec:vlms}

In recent years, vision-language models have developed rapidly, aiming to enhance the generalization ability of vision models to unseen object categories. The core idea is to leverage large-scale image-text pairs and train networks via noise-contrastive learning, enabling the alignment of image representations with text embeddings. Recent studies, represented by CLIP~\cite{CLIP2021ICML} and ALIGN~\cite{align2021icml}, use millions of image-text pairs to augment the training process and adopt Transformers as the backbone network. Their rich vision-language correspondence knowledge serves as effective pretrained models, applicable to various tasks ranging from few-shot to zero-shot settings, such as image captioning~\cite{vltint2023aaai} and semantic segmentation~\cite{decouplingding2022cvpr, aisformer2022arxiv}. Meanwhile, adaptation methods for large-scale vision-language models have become a popular research direction; these methods aim to perform minimal fine-tuning on these computationally intensive models while enhancing their generalization ability on new tasks. For instance, some previous works on OV-TAL have explored adaptation methods~\cite{effprompt2022eccv, stable2022eccv, zeetad2024wacv} such as text prompt tuning~\cite{lester2021arxiv} to apply these models to downstream tasks. In this work, we utilize the text encoder of CLIP to directly extract text features without employing any text prompt fine-tuning.

\begin{figure*}[h]
  \centering
  \includegraphics[width=1.0\linewidth]{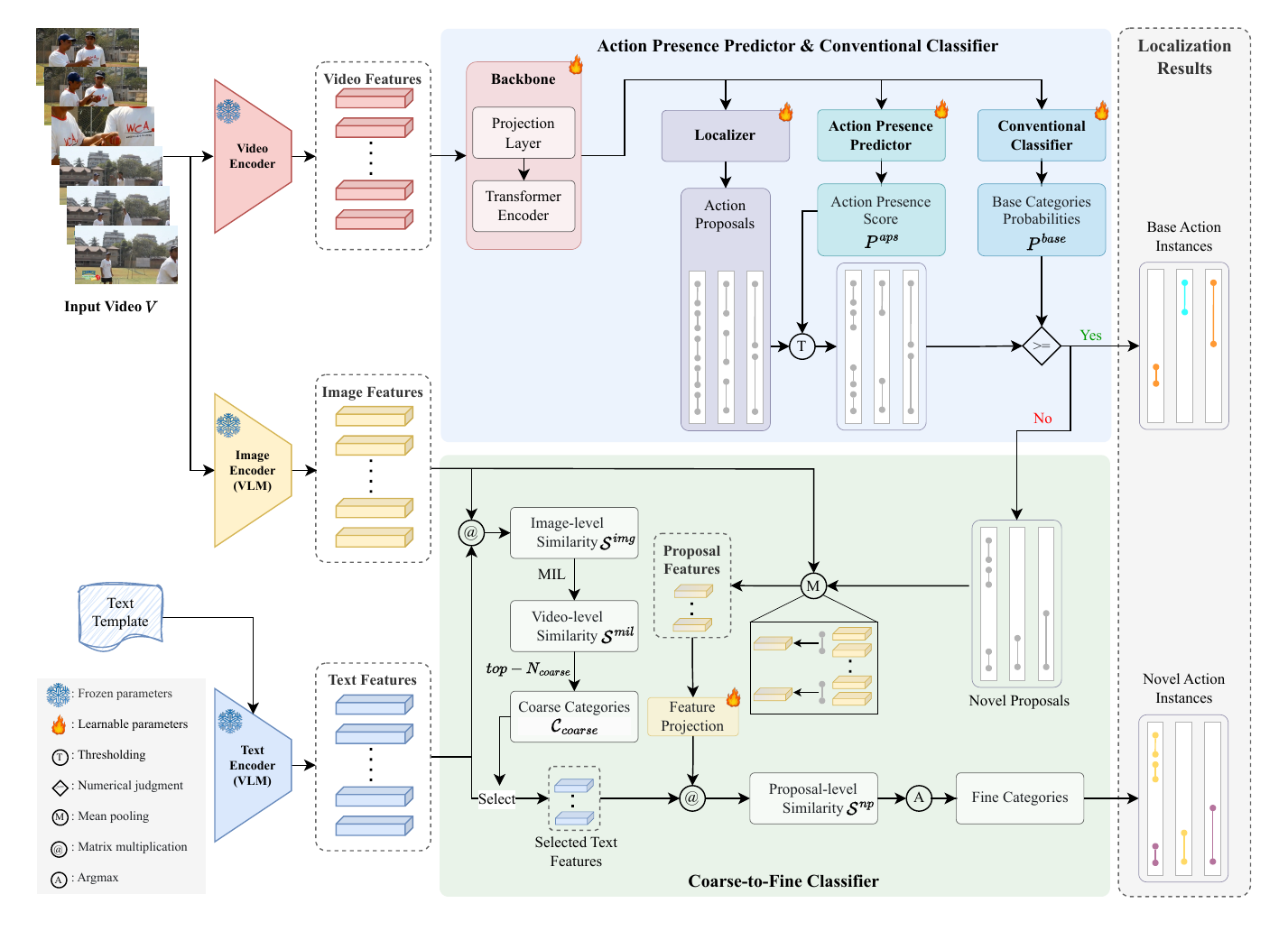}
  \caption{Overview of the proposed MGCA-Net. MGCA-Net employs frozen video, image, and text encoders to extract video, image, and text features. Based on video features, the localizer localizes category-agnostic action proposals. In parallel, the action presence predictor and conventional classifier predict the action presence score (APS) for each action proposal and its probabilities over base action categories. Based on the APS and base action probabilities, action proposals are categorized into base action instances, novel proposals, or discarded. The coarse-to-fine classifier determines the action categories of novel proposals. Specifically, the coarse-to-fine classifier first identifies all action categories in the video, yielding coarse categories. Subsequently, it extracts proposal features for each novel proposal and assigns an action category based on the similarity between the proposal features and the text features of the coarse categories, resulting in novel action instances. The final localization results are the union of base and novel action instances.}
  \label{fig:method_main}
\end{figure*}

\subsection{Temporal Action Localization}\label{subsec:tal}

TAL is a fundamental problem in video understanding. Most localization frameworks can be categorized into two groups: two-stage~\cite{rc3d2017CVPR, pcpcad2019, gtad2020CVPR, tallformer2022ECCV} \textit{v.s.} one-stage~\cite{actionformer2022, tridet2023, bdrcnet2025} detectors. These localizers typically involve several heuristic steps, such as thresholding and non-maximum suppression (NMS). Recently, TadTR~\cite{tadtr2021tip}, an end-to-end localizer for TAL task based on transformer~\cite{transformer2017NeurIPS}, has been proposed. Similar to DETR~\cite{detr2020ECCV}, it formulates localization as a set-to-set prediction problem, eliminating some of the previous heuristics and enabling a simpler localization pipeline. However, the aforementioned methods rely on large amounts of annotated data, limiting their practical applications. Thus, weakly supervised TAL has been proposed. Weakly supervised TAL only relies on video category annotations~\cite{STPN2018CVPR, pivotaltal2023CVPR, gaufuse2023CVPR} or point annotations~\cite{LACP2021ICCV, lpr2023MMSJ, HRPro2024AAAI, SMBD2024ECCV}, reducing the cost of data annotation. However, these methods can only predict action categories within a closed set and fail to localize novel action categories during inference, while OV-TAL fills this gap.

\subsection{Open-Vocabulary Temporal Action Localization}\label{subsec:ov_tal}

OV-TAL extends traditional TAL to open-vocabulary scenarios. Existing OV-TAL methods can be broadly categorized into two paradigms: localization-then-classification and text information injection. The localization-then-classification methods~\cite{effprompt2022eccv, detal2024tpami, zeetad2024wacv, stovtal2025wacv} first generate category-agnostic temporal proposals, then extract proposal-level features and leverage the zero-shot capabilities of VLMs for action classification. Most approaches in this category explore adapting pre-trained VLMs to downstream OV-TAL tasks via text prompt tuning. In contrast, text information injection methods~\cite{unloc2023iccv, mprotea2024tcsvt, tifad2024neurips, ovformer2024bmvc} typically freeze the text encoder of VLMs and inject textual knowledge into the localization model through cross-attention or similar mechanisms, enabling the detection of novel action categories by aligning visual and textual representations.

However, these methods all perceive action categories at a single granularity, which restricts their capability to perceive action categories. Our method addresses this limitation by adopting a multi-grained category-aware framework that decouples the prediction of base and novel actions.

\section{Method}\label{sec:method}

\subsection{Problem Definition}\label{subsec:probledefinition}

Given a video $V$ in the training set, its corresponding annotations are denoted as $\Psi = \{\psi_i = (t_{s,i}, t_{e,i}, c_i)\}_{i=1}^{N_a}$, where $N_a$ is the number of action instances, $t_{s,i}$ and $t_{e,i}$ represent the start and end times of the $i$-th action instance $\psi_i$, respectively. $c_i \in \mathcal{C}_{\text{base}}$ is the action category of the $\psi_i$, and $\mathcal{C}_{\text{base}}$ denotes the set of base action categories, i.e., all annotated action categories in the training set. The objective of OV-TAL is to train a TAL model on the training set, which can predict action instances corresponding to all desired action categories $\mathcal{C}_{\text{all}}$ during inference. Formally, $\mathcal{C}_{\text{all}} = \mathcal{C}_{\text{base}} \cup \mathcal{C}_{\text{novel}}$ with $\mathcal{C}_{\text{base}} \cap \mathcal{C}_{\text{novel}} = \emptyset$, where $\mathcal{C}_{\text{novel}}$ denotes the set of novel action categories that are expected to be predicted during inference.

\subsection{Overview}\label{subsec:oveview}

As shown in Fig.~\ref{fig:method_main}, our proposed MGCA-Net comprises a localizer, an action presence predictor, a conventional classifier, and a coarse-to-fine classifier. Given an input video $V$, following existing methods~\cite{effprompt2022eccv, detal2024tpami, zeetad2024wacv}, MGCA-Net extracts video features $F^{vid} \in \mathbb{R}^{T_{vid} \times D_{vid}}$ and image features $F^{img} \in \mathbb{R}^{T_{img} \times D_{img}}$ using a pre-trained video encoder and the image encoder of VLMs, respectively. Where, $T_{\text{vid}}$, $D_{\text{vid}}$, $T_{\text{img}}$, and $D_{\text{img}}$ denote the temporal length and dimension of the features $F^{\text{vid}}$ and $F^{\text{img}}$, respectively. Meanwhile, for text descriptions of action categories, MGCA-Net constructs a fixed text template to extract text features $F^{text} \in \mathbb{R}^{|\mathcal{C}_{novel}| \times D_{text}}$ via the text encoder of VLMs. Where, $|\mathcal{C}_{novel}|$ and $D_{\text{text}}$ denote the number of $\mathcal{C}_{novel}$ and dimension of the feature $F^{\text{text}}$, respectively. Next, the backbone network takes video features as input to model temporal context information, and predicts category-agnostic action proposals through the localizer. For each action proposal, the action presence predictor and conventional classifier respectively predict their corresponding action presence score (APS) and probabilities of base actions. Based on the APS and base action probabilities, each action proposal is determined to belong to either a base or a novel proposal. Specifically, proposals where both the APS and base action probabilities exceed the threshold are categorized as base proposals. Otherwise, if the APS exceeds the threshold, the proposals are categorized as novel proposals. The remaining proposals are regarded as noise and discarded.

The base action category with the highest base action probability determines the base proposals' action category. For novel proposals, their action category is determined by the coarse-to-fine classifier of MGCA-Net. Specifically, the coarse-to-fine classifier utilizes image and text features to measure the similarity between each image and the text descriptions of actions, and employs a multi-instance learning (MIL) algorithm to identify all action categories occurring in video $V$, referred to as coarse categories. For each novel proposal, the average of its corresponding image features is extracted as the proposal feature. Finally, it computes the similarity between the proposal feature and the text features corresponding to the coarse categories; the action category associated with the maximum similarity is designated as the category of this proposal, i.e., the fine category.

\subsection{Localizer and Conventional Classifier}\label{subsec:localizer}

The localizer and conventional classifier aim to localize category-agnostic action proposals in the input video and predict the probability of each proposal belonging to each base action category, respectively. They are widely used in TAL. In MGCA-Net, we employ the localizer and conventional classifier from ActionFormer~\cite{actionformer2022}, whose localizer has been proven effective in existing OV-TAL methods~\cite{effprompt2022eccv, detal2024tpami, tifad2024neurips}.

\textbf{Backbone.} As shown in Fig.~\ref{fig:method_main}, given video features $F^{\text{vid}}$, the transformer backbone of ActionFormer is first used to model temporal context features, yielding $L$-level FPN features $F^{\text{fpn}} \in \mathbb{R}^{T_{\text{fpn}} \times D_{\text{fpn}}}$, where $ T_{\text{fpn}} $ and $ D_{\text{fpn}} $ denote the feature length and dimension of $ F^{\text{fpn}} $, respectively. Subsequently, $ F^{\text{fpn}} $ is fed as input to the localizer, conventional classifier, and action presence predictor.

\textbf{Localizer.} Based on $ F^{\text{fpn}} $, the localizer first models features suitable for localization using two 1D convolutions (conv1d) with kernel size 3, stride 1, and padding 1. Subsequently, it predicts onset and offset pairs $ \{(d_{\text{on}, i}, d_{\text{off},i})\}_{i=1}^{T_{\text{fpn}}} $ at each temporal position via a conv1d with kernel size 1, stride 1, and padding 1, where $ (d_{\text{on},i}, d_{\text{off},i}) $ denote the predicted onset and offset corresponding to the $ i $-th temporal position $ t_i $, respectively. The action proposal corresponding to the $ i $-th temporal position is decoded as $ \psi^p_i = (t^p_{s,i},t^p_{e,i}) = (t_i - d_{\text{on},i}, t_i + d_{\text{off},i}) $, which is subsequently re-scaled based on specific FPN scales. Finally, this yields the set of action proposals, denoted as $\Psi^p = \{\psi^p_{i}\}_{i=1}^{T_{\text{fpn}}}$.

\textbf{Conventional Classifier.} The conventional classifier first models features using two conv1d layers with the same parameters as the localizer, then predicts the probabilities of each action proposal belonging to base action categories via a conv1d layer with an output dimension of $|\mathcal{C}_{\text{base}}|$, resulting in $P^{\text{base}} \in \mathbb{R}^{T_{\text{fpn}} \times |\mathcal{C}_{\text{base}}|}$, where $|\mathcal{C}_{\text{base}}|$ denotes the number of $\mathcal{C}_{\text{base}}$.

\textbf{Loss Function.} During training, the localizer and conventional classifier utilize the same loss functions as their corresponding modules in ActionFormer. For the localizer, it employs DIoU loss~\cite{diou2020aaai} to compute the regression loss $\mathcal{L}_{\text{loc}}$ between the predicted action proposals and their corresponding ground truth. The conventional classifier employs focal loss~\cite{focalloss2017ICCV} to compute the classification loss $\mathcal{L}_{\text{cc}}$ between the predicted probabilities and their corresponding ground truth.

\begin{algorithm}
    \caption{Ground truth generation for action presence score}
    \label{algo:aps_groundtruth}
    \parbox[t]{\linewidth}{\textbf{Input:}}\hspace{0pt}
        \begin{algorithmic}[0]
          \State Annotations $\Psi = \{\psi_i = (t_{s,i}, t_{e,i}, c_i)\}_{i=1}^{N_a}$; Action proposals $\Psi^p=\{\psi^p_i\}_{i=1}^{T_{fpn}}$.
        \end{algorithmic}
    \parbox[t]{\linewidth}{\textbf{Output:}}\hspace{0pt}
        \begin{algorithmic}[0]
          \State Ground truth for action presence score $\hat{P}^{aps}$.
        \end{algorithmic}

        \begin{algorithmic}[1]
            \Statex

            \State Initialize $P^{loc}$ and $\hat{P}^{aps}$ as $T_{fpn}$-length zero vectors: $P^{loc}, \hat{P}^{aps} \gets \mathbf{0}^{T_{fpn}}$
            \State{$P^{gt} \gets \{\}$} \Comment{Initialize empty set}
            
            \Statex
            \For{$i \gets 1$ \textbf{to} $T_{fpn}$}
                \For{$j \gets 1$ \textbf{to} $N_{a}$}
                    \If{$t_{s,j} \le t_i$ \textbf{and} $t_i \le t_{e,j}$}
                        \State $P^{loc}_i \gets 1$
                        \State add $(t_{s,j}, t_{e,j})$ to $P^{gt}$
                    \Else
                        \State add $(-1, -1)$ to $P^{gt}$
                    \EndIf
                \EndFor
            \EndFor
            \Statex

            \For{$i \gets 1$ \textbf{to} $T_{fpn}$}
                \If{$P^{loc}_i == 1$}
                    \State $\hat{P}^{aps}_i \gets tIoU(\psi^p_i, P^{gt}_i)$
                \EndIf
            \EndFor
            \Statex
            
            \State \textbf{return} $\hat{P}^{aps}$

        \end{algorithmic}
\end{algorithm}

\subsection{Action Presence Predictor}\label{subsec:app}

The action presence predictor aims to estimate the probability of each action proposal being an action instance. Taking the features $ F^{\text{fpn}} $ as input, it operates in parallel with the localizer and conventional classifier. Specifically, it first employs two conv1d with a kernel size of 3, a stride of 1, and a padding of 1 to model features. Subsequently, a conv1d layer with a kernel size, stride, and padding of 1, and an output dimension of 1, is used to predict the action presence probability at each temporal position, yielding the action presence score (APS) $ P^{\text{aps}} \in \mathbb{R}^{T_{\text{fpn}}} $. $P^{\text{aps}}_{i}$ denotes the probability that the $i$-th action proposal $\psi^p_{i}$ is an action instance.

\begin{algorithm}
    \caption{Generate base action instances and novel proposals}
    \label{algo:generate_bnp}
    \parbox[t]{\linewidth}{\textbf{Input:}}\hspace{0pt}
        \begin{algorithmic}[0]
          \State Action proposal $\Psi^p = \{\psi^p_i = (t^p_{s,i}, t^p_{e,i})\}_{i=1}^{T_{\text{fpn}}}$; Action presence scores $P^{\text{aps}} \in \mathbb{R}^{T_{\text{fpn}}}$; Base action probabilities $P^{\text{base}} \in \mathbb{R}^{T_{\text{fpn}} \times |\mathcal{C}_{\text{base}}|}$; Threshold $\lambda_{retain}$ and $\lambda_{base}$.
        \end{algorithmic}
    \parbox[t]{\linewidth}{\textbf{Output:}}\hspace{0pt}
        \begin{algorithmic}[0]
          \State Base action instances $\Psi^{base}$; Novel proposals $\Psi^{np}$.
        \end{algorithmic}

        \begin{algorithmic}[1]
            \Statex

            \State{$\Psi^{base} \gets \{\}$} \Comment{Initialize empty set}
            \State{$\Psi^{np} \gets \{\}$} \Comment{Initialize empty set}

            \Statex
            \For{$i \gets 1$ \textbf{to} $T_{fpn}$}
                \If{$P^{aps}_i \ge \lambda_{retain}$ \textbf{and} $max(P^{base}_i) \ge \lambda_{base}$}
                    \State add $\{t^p_{s,i}, t^p_{e,i}, \mathcal{C}_{base}[argmax(P^{base}_i)]\}$ to $\Psi^{base}$
                \ElsIf{$P^{aps}_i \ge \lambda_{retain}$}
                    \State add $\psi^{p}_{i}$ to $\Psi^{np}$
                \EndIf
            \EndFor
            \Statex
            
            \State \textbf{return} $\Psi^{base}$, $\Psi^{np}$.

        \end{algorithmic}
\end{algorithm}

To avoid interference from category information of base actions, during training, we use the temporal intersection over union (tIoU) between each action proposal and its corresponding action instance as the ground truth for $P^{\text{aps}}$. Let the ground truth be denoted as $\hat{P}^{\text{aps}}$, whose construction process is detailed in Algo.~\ref{algo:aps_groundtruth}. First, we determine positive temporal locations based on the annotation information $\{(t_{s,i}, t_{e,i}, c_i)\}_{i=1}^{N_a}$. Specifically, we initialize $P^{\text{loc}} \in \mathbb{R}^{T_{\text{fpn}}}$ to 0, where $P^{\text{loc}}_i$ denotes the value corresponding to the $i$-th temporal position $t_i$. For a temporal position $t_i$, if it lies within a labeled temporal interval (e.g., for the $j$-th annotation, $t_{s,j} \le t_i \le t_{e,j}$), we set $P^{\text{loc}}_i$ to 1. Subsequently, we compute $\hat{P}^{\text{aps}}$ based on $P^{\text{loc}}$. Specifically, all values of $\hat{P}^{\text{aps}}$ are also initialized to 0. For a temporal position $t_i$, if $P^{\text{loc}}_i = 1$, we compute the tIoU between the proposal $\psi^p_i$ at this position and its corresponding ground truth, and take this tIoU as the value of $\hat{P}^{\text{aps}}_i$. This construction of $\hat{P}^{\text{aps}}$ effectively avoids its dependence on classification information, enabling the action presence score to more effectively focus on the localization quality of the proposal itself.

\textbf{Loss Function.} Given the prediction of APS $ P^{\text{aps}} $ and their corresponding ground truth $ \hat{P}^{\text{aps}} $, we employ the $\mathcal{L}_1$ loss as the loss function, where the loss $ \mathcal{L}_{\text{app}} $ is computed as shown in Eq.~\ref{eq:loss_app}.

\begin{equation}\label{eq:loss_app}
  \mathcal{L}_{app} = \frac{1}{\sum_{i=1}^{T_{fpn}} P^{loc}_i} \sum_{i=1}^{T_{fpn}} \mathbb{I}(P^{loc}_i) \mathcal{L}_1(P^{\text{aps}}_i, \hat{P}^{\text{aps}}_i)
\end{equation}

where $\mathbb{I}(\cdot)$ is the indicator function.

\subsection{Identify Base Action Instances or Novel Proposals}\label{subsec:identitybaseornovel}

Given the action proposals $\Psi^p = \{\psi^p_i = (t^p_{s,i}, t^p_{e,i})\}_{i=1}^{T_{\text{fpn}}}$ output by the localizer, the APS $P^{\text{aps}} \in \mathbb{R}^{T_{\text{fpn}}}$ predicted by the action presence predictor, and the probabilities $P^{\text{base}} \in \mathbb{R}^{T_{\text{fpn}} \times |\mathcal{C}_{\text{base}}|}$ predicted by the conventional classifier, we need to categorize these action proposals into base action instances $\Psi^{base}$, novel proposals $\Psi^{np}$, or discarded ones.

Specifically, as shown in Algo.~\ref{algo:generate_bnp}, for the $ i $-th proposal $ \psi^p_i $ in $ \Psi^p $, its corresponding action presence score and base action probabilities are respectively $ P^{\text{aps}}_i \in \mathbb{R}^1 $ and $ P^{\text{base}}_i \in \mathbb{R}^{|\mathcal{C}_{\text{base}}|} $. We first check whether the conditions $ P^{\text{aps}}_i \ge \lambda_{\text{retain}} $ and $ \max(P^{\text{base}}_i) \ge \lambda_{\text{base}} $ are both satisfied. If both conditions are met, the proposal is added to the set of base action instances $\Psi^{base}$, with its category being $ \mathcal{C}_{\text{base}}[\arg\max(P^{\text{base}}_i)] $, and its start and end times being $ t^p_{s,i} $ and $ t^p_{e,i} $, respectively. If only $ P^{\text{aps}}_i \ge \lambda_{\text{retain}} $ is satisfied, $ \psi^p_i $ is added to the set of novel proposals $ \Psi^{\text{np}} $. The remaining action proposals will be discarded.

Finally, we obtain the prediction results for base action categories, denoted as $\Psi^{\text{base}} = \{\psi^{\text{base}}_i = (t^{\text{base}}_{s,i}, t^{\text{base}}_{e,i}, c^{\text{base}}_i)\}_{i=1}^{N_{\text{base}}}$, and the novel proposals $\Psi^{\text{np}} = \{\psi^{\text{np}}_i = (t^{\text{np}}_{s,i}, t^{\text{np}}_{e,i})\}_{i=1}^{N_{\text{np}}}$. Here, $t^{\text{base}}_{s,i}$, $t^{\text{base}}_{e,i}$, and $c^{\text{base}}_i$ denote the start time, end time, and action category of the $i$-th base action instance $\psi^{\text{base}}_i$, respectively; $t^{\text{np}}_{s,i}$ and $t^{\text{np}}_{e,i}$ denote the start time and end time of the $i$-th novel proposal $\psi^{\text{np}}_i$, respectively.

\subsection{Coarse-to-Fine Classifier}\label{subsec:ctfc}

The coarse-to-fine classifier aims to leverage VLMs' zero-shot capability to identify each novel proposal's action category in $\Psi^{\text{np}}$. As shown in Fig.~\ref{fig:method_main}, we first address the problem of action presence—i.e., whether an action category is present in the input video—yielding coarse categories. Subsequently, we extract proposal-level features using the temporal intervals of novel proposals and image features $F^{\text{img}}$. We then determine which action category within the coarse categories each novel proposal belongs to, based on the proposal-level features and text features $F^{\text{text}}$, yielding fine categories.

\textbf{Generate Coarse Categories.} To identify the action categories present in the input video, MGCA-Net obtains coarse categories by computing the similarity between images and action texts. Given the image features $ F^{\text{img}} $ of the input video and the text features $ F^{\text{text}} $ of action categories, we first calculate the similarity between each image feature and all action categories in $ \mathcal{C}_{\text{novel}} $, resulting in $ \mathcal{S}^{\text{img}} \in \mathbb{R}^{T_{img} \times |\mathcal{C}_{novel}|}$, as shown in Eq.~\ref{eq:simg}.

\begin{equation}\label{eq:simg}
  \mathcal{S}^{img} = F_{img} \cdot transpose(F^{text})
\end{equation}

where $\cdot$ denotes matrix multiplication, and $transpose()$ denotes the transpose of a matrix. $\mathcal{S}^{\text{img}}_i \in \mathbb{R}^{|\mathcal{C}_{\text{novel}}|}$ denotes the $i$-th element of $\mathcal{S}^{\text{img}}$, which represents the similarity between the $i$-th image and each novel action in $\mathcal{C}_{\text{novel}}$.

Subsequently, based on $\mathcal{S}^{\text{img}}$, we compute the similarity between the entire input video and each novel action via multi-instance learning (MIL), yielding $\mathcal{S}^{\text{mil}} \in \mathbb{R}^{|\mathcal{C}_{\text{novel}}|}$. Specifically, the probability of the $k$-th novel action category is denoted as $\mathcal{S}^{\text{mil}}_k$,
it is calculated using the Eq.~\ref{eq:smil}.

\begin{equation}\label{eq:smil}
  \mathcal{S}^{mil}_k = \max_{\substack{H \subset \mathcal{S}^{img}_{:, k} \\ |H| = T_{img}/8}} \frac{1}{|H|} \sum_{h=1}^{|H|}H_h
\end{equation}

where $\mathcal{S}^{\text{img}}_{:, k}$ denotes the probabilities of the $k$-th action category across all images in $\mathcal{S}^{\text{img}}$, and $H$ consists of the top-$T_{\text{img}}/8$ values in $\mathcal{S}^{\text{img}}_{:, k}$, with $H_h$ denoting the $h$-th value in $H$. Within $\mathcal{S}^{\text{mil}}$, the categories corresponding to the top-$N_{\text{coarse}}$ values form the coarse categories $\mathcal{C}_{\text{coarse}}$.

\textbf{Multiple Templates Fusion.} When the text encoder is frozen, we use text templates to extract text features of action categories, such as \textit{the action of \{action name\}.} However, a single text template may fail to capture all features of the entire action. Therefore, inspired by the success of multi-template in open-vocabulary object detection~\cite{vild2022iclr, ovdquo2025aaai}, we use multiple templates to extract diverse text features and take the average of all text features as $F^{\text{text}}$.

\textbf{Generate Fine Categories.} Coarse categories $\mathcal{C}_{\text{coarse}}$ are only capable of identifying all action categories present in the input video. However, they cannot determine which category each novel proposal in $\Psi^{\text{np}}$ belongs to. Thus, we need to assign action categories to these novel proposals further.

Specifically, we first extract proposal features $F^{\text{np}}$. For the $ i $-th novel proposal $ \psi^{\text{np}}_i $, its proposal feature $ F^{\text{np}}_i $ is obtained from Eq.~\ref{eq:fnpi}.

\begin{equation}\label{eq:fnpi}
  F^{\text{np}}_i = mean(F^{img}_{t^{np}_{s,i}:t^{np}_{e,i}, :})
\end{equation}

where $mean(\cdot)$ denotes the average operation, and $F^{\text{img}}_{t^{\text{np}}_{s,i}:t^{\text{np}}_{e,i}, :}$ denotes the features of $F^{\text{img}}$ within the temporal interval $[t^{\text{np}}_{s,i}, t^{\text{np}}_{e,i}]$. Subsequently, we use a feature projection layer $\phi_{\text{proj}}$ to align proposal features $F^{\text{np}}$ with text features. $\phi_{\text{proj}}$ consists of two linear layers, whose input and output dimensions are consistent with the feature dimension of $F^{\text{np}}$.

Meanwhile, based on the text features $F^{\text{text}}$, we select the corresponding feature for each category in $\mathcal{C}_{\text{coarse}}$, obtaining the features of coarse categories denoted as $F^{\text{coarse}} \in \mathbb{R}^{|\mathcal{C}_{\text{coarse}}| \times D_{\text{text}}}$. Then, $\mathcal{S}^{\text{np}} \in \mathbb{R}^{N_{\text{np}} \times |\mathcal{C}_{\text{coarse}}|}$ is computed via Eq.~\ref{eq:snp}, where $\mathcal{S}^{\text{np}}_i$ represents the similarity between the $i$-th novel proposal $\psi^{\text{np}}_i$ in $\Psi^{\text{np}}$ and the coarse categories.

\begin{equation}\label{eq:snp}
  \mathcal{S}^{np} = F^{np} \cdot transpose(F^{coarse})
\end{equation}

where $\cdot$ denotes matrix multiplication, and $transpose()$ denotes the transpose of a matrix.

\textbf{Generate Novel Action Instance.} Given the novel proposals $\Psi^{\text{np}}$ and similarities $\mathcal{S}^{\text{np}}$, we determine the category of each novel proposal as the category corresponding to the maximum similarity among all coarse action categories, yielding the novel action instances $\Psi^{\text{novel}} = \{\psi^{\text{novel}}_i\}_{i=1}^{N_{\text{np}}}$. Specifically, for the $i$-th novel proposal $\psi^{\text{np}}_i$, its corresponding action instance is $\psi^{\text{novel}}_i = (t^{\text{np}}_{s,i}, t^{\text{np}}_{e,i}, c^{\text{np}}_i)$, where $c^{\text{np}}_i$ is its corresponding action category, determined by the action category in $\mathcal{C}_{\text{coarse}}$ that corresponds to the maximum value of $\mathcal{S}^{\text{np}}_i$.

\textbf{Loss Function.} For the generation of coarse categories, our MGCA-Net adopts a training-free manner, which is based on features extracted from frozen pre-trained VLMs and MIL. Thus, there are no parameters requiring optimization in this sub-module, and no loss function needs to be computed. For generating fine categories, it is necessary to compute a loss to train the projection layer $\phi_{\text{proj}}$, which is implemented by calculating the contrastive loss $\mathcal{L}_{\text{contrast}}$.

Specifically, during training, since only annotations corresponding to base action categories are available, we directly set the content of $\mathcal{C}_{\text{novel}}$ to be identical to $\mathcal{C}_{\text{base}}$. For the proposal feature $F^{\text{np}}_i$ corresponding to the $i$-th novel proposal $\psi^{\text{np}}_i$ in $\Psi^{\text{np}}$, we take the text feature of its corresponding ground-truth action category as the positive feature $F^{\text{pos}}_i \in \mathbb{R}^{1 \times D_{\text{img}}}$. Furthermore, we randomly select text features corresponding to $N_{\text{neg}}$ other categories from $\mathcal{C}_{\text{novel}}$ as negative features $F^{\text{neg}}_i \in \mathbb{R}^{N_{\text{neg}} \times D_{\text{img}}}$. By merging $F^{\text{pos}}_i$ and $F^{\text{neg}}_i$, we obtain the contrastive feature $F^{\text{contrast}}_i = concatenate((F^{\text{pos}}_i, F^{\text{neg}}_i), dim=0)$, where $F^{\text{contrast}}_i \in \mathbb{R}^{(N_{\text{neg}} + 1) \times D_{\text{img}}}$, and $concatenate((\cdot), dim=0)$ denotes concatenation along the first dimension. After extracting contrastive features for all novel proposals, we obtain $F^{\text{contrast}}$. Based on Eq.~\ref{eq:snp}, the similarities $\mathcal{S}^{\text{np}} \in \mathbb{R}^{N_{\text{np}} \times (N_{\text{neg}} + 1)}$ can be calculated. Their corresponding labels are all the first element, i.e., $\hat{\mathcal{S}}^{\text{np}} = \text{zeros}(N_{\text{np}})$. Finally, $\mathcal{L}_{\text{contrast}}$ is computed using the cross-entropy loss as shown in Eq.~\ref{eq:conloss}.

\begin{equation}\label{eq:conloss}
  \mathcal{L}_{\text{contrast}} = CrossEntropyLoss(\mathcal{S}^{\text{np}}, \hat{\mathcal{S}}^{\text{np}})
\end{equation}

\subsection{Training and Inference}\label{subsec:train_infer}

\textbf{Training.} During training, MGCA-Net jointly trains all modules based on the total loss function $\mathcal{L}$.

\begin{equation}\label{eq:fullloss}
  \mathcal{L} = \mathcal{L}_{\text{loc}} + \mathcal{L}_{\text{cc}} + \mathcal{L}_{\text{app}} + \mathcal{L}_{\text{contrast}}
\end{equation}

\textbf{Inference.} During inference, MGCA-Net first determines the prediction results of base actions, $\Psi^{\text{base}}$, via Sec.~\ref{subsec:identitybaseornovel}. Subsequently, it determines the prediction results of novel actions, $\Psi^{\text{novel}}$, via Sec.~\ref{subsec:ctfc}. The final prediction results $\Psi^{\text{all}}$ are obtained as the union of $\Psi^{\text{base}}$ and $\Psi^{\text{novel}}$.

\section{Experiments}\label{sec:experiments}

\begin{table*}[t!]
    \centering
    \caption{Performance comparison of OV-TAL with state-of-the-art methods on THUMOS'14 and ActivityNet-1.3. mAP$_{\text{base}}$, mAP$_{\text{novel}}$, and mAP$_{\text{all}}$ denote the average of mAP across different tIoU thresholds on base, novel, and all action categories for different datasets, respectively. \textbf{Bold} values indicate the best results, and \underline{underlined} values indicate the second-best results. $^*$ denotes the reproduced results of OVFormer.}
    \resizebox{0.90\linewidth}{!}{\begin{tabular}{cl|cccccc}
\toprule
\multirow{2}{*}{\textbf{Split}} & \multicolumn{1}{c|}{\multirow{2}{*}{\textbf{Method}}} & \multicolumn{3}{c}{\textbf{THUMOS'14}} & \multicolumn{3}{c}{\textbf{ActivityNet-1.3}} \\ \cmidrule{3-8}
 & \multicolumn{1}{c|}{} & \textbf{mAP$_{base}$} & \textbf{mAP$_{novel}$} & \multicolumn{1}{c|}{\textbf{mAP$_{all}$}} & \textbf{mAP$_{base}$} & \textbf{mAP$_{novel}$} & \textbf{mAP$_{all}$} \\ \midrule
\multirow{6}{*}{\begin{tabular}[c]{@{}c@{}}75\% Seen\\ 25\% Unseen\end{tabular}} & P-ActionFormer~\cite{ovformer2024bmvc} & 51.9 & 13.8 & \multicolumn{1}{c|}{41.5} & 30.0 & 15.3 & 26.3 \\
 & L-ActionFormer~\cite{ovformer2024bmvc} & 52.3 & 14.7 & \multicolumn{1}{c|}{42.8} & 30.9 & 16.8 & 27.3 \\
 & F-ActionFormer~\cite{ovformer2024bmvc} & 50.8 & 24.2 & \multicolumn{1}{c|}{44.1} & 30.8 & 22.9 & 28.8 \\
 & STABLE$^*$~\cite{stable2022eccv} & - & - & \multicolumn{1}{c|}{-} & 23.2 & 20.6 & 22.6 \\
 & OVFormer~\cite{ovformer2024bmvc} & \underline{56.4} & \underline{27.3} & \multicolumn{1}{c|}{\underline{49.1}} & \underline{31.4} & \underline{25.1} & \underline{29.8} \\
 & MGCA-Net (Ours) & \textbf{59.2} & \textbf{34.0} & \multicolumn{1}{c|}{\textbf{52.9}} & \textbf{31.6} & \textbf{28.6} & \textbf{30.8} \\ \midrule
\multirow{6}{*}{\begin{tabular}[c]{@{}c@{}}50\% Seen\\ 50\% Unseen\end{tabular}} & P-ActionFormer~\cite{ovformer2024bmvc} & 50.9 & 9.9 & \multicolumn{1}{c|}{30.5} & 27.6 & 13.0 & 20.3 \\
 & L-ActionFormer~\cite{ovformer2024bmvc} & 48.3 & 10.1 & \multicolumn{1}{c|}{29.2} & 28.3 & 13.5 & 20.9 \\
 & F-ActionFormer~\cite{ovformer2024bmvc} & 51.2 & 20.5 & \multicolumn{1}{c|}{35.8} & 28.8 & 23.5 & 26.2 \\
 & STABLE$^*$~\cite{stable2022eccv} & - & - & \multicolumn{1}{c|}{-} & 23.0 & 20.7 & 22.2 \\
 & OVFormer~\cite{ovformer2024bmvc} & \underline{55.7} & \underline{24.9} & \multicolumn{1}{c|}{\underline{40.7}} & \underline{30.2} & \underline{24.8} & \underline{27.5} \\
 & MGCA-Net (Ours) & \textbf{56.3} & \textbf{31.3} & \multicolumn{1}{c|}{\textbf{43.9}} & \textbf{30.3} & \textbf{28.2} & \textbf{29.2} \\ \bottomrule
\end{tabular}}
\label{tab:ov_main}
\end{table*}

\subsection{Experiment Setting}\label{subsec:expsetting}

\textbf{Datasets.} We conduct experiments on THUMOS'14~\cite{thumos142017CVIU} and ActivityNet-1.3~\cite{activitynet2015CVPR}, two benchmark datasets commonly used in TAL and OV-TAL. THUMOS'14 contains 20 sports action categories, with 200 training videos and 213 test videos; each video typically includes multiple distinct action instances, increasing this dataset's difficulty. ActivityNet-1.3 comprises 200 daily life action categories, with 19,994 videos. Following the standard setup~\cite{effprompt2022eccv}, we split the dataset into training, validation, and test sets at a ratio of 2:1:1.

In the open-vocabulary setting, as shown in~\cite{effprompt2022eccv}, we set two scenarios: the 50\%-50\% setup (50\% of action categories are used for training, and the remaining for testing) and the 75\%-25\% setup (75\% of action categories are used for training, and the remaining for testing). Additionally, we perform 10 random splits to make it statistically robust and take the average over the final performance.

\textbf{Evaluation Metric.} Following the standard evaluation protocol, we report mean Average Precision (mAP) at various tIoU thresholds. Specifically, for the THUMOS'14 dataset, we set the tIoU thresholds from 0.3 to 0.7 with an interval of 0.1 (i.e., [0.3:0.1:0.7]); for the ActivityNet-1.3 dataset, the tIoU thresholds are set from 0.5 to 0.95 with an interval of 0.05 (i.e., [0.5:0.05:0.95]). For base, novel, and all action categories, we measure the corresponding metrics mAP$_{\text{base}}$, mAP$_{\text{novel}}$, and mAP$_{\text{all}}$, respectively.

\textbf{Implementation Details.} Following existing methods~\cite{zeetad2024wacv, detal2024tpami}, we adopt I3D~\cite{i3dkinetics2017CVPR} and TSP~\cite{tsp2021iccvw}, both pre-trained on the Kinetics~\cite{kinetics2017arxiv} dataset, as the video encoder of MGCA-Net for THUMOS'14 and ActivityNet-1.3, respectively. Meanwhile, we use a frozen pre-trained CLIP (ViT-B/16)~\cite{CLIP2021ICML} as the image and text encoder. This model is widely used in existing methods~\cite{tifad2024neurips, mprotea2024tcsvt, zeetad2024wacv}, which ensures the fairness of performance comparison. Our model is trained for 35 epochs on THUMOS'14 and 15 epochs on ActivityNet-1.3, using AdamW with 5 epochs of linear warmup. The initial learning rates for THUMOS'14 and ActivityNet-1.3 are set to 1e-4 and 1e-3, respectively, both updated with cosine annealing~\cite{loshchilov2016sgdr}. The hyperparameters $\lambda_{\text{retain}}$ and $\lambda_{\text{base}}$ are both set to 0.5, and $N_{\text{coarse}}$ is set to 2. When training $\phi_{\text{proj}}$, $N_{\text{neg}}$ is set to 3 to balance the ratio of positive to negative samples at 1:3. All experiments are conducted with a single NVIDIA RTX 3090 GPU.

\subsection{Open-Vocabulary Results}\label{subsec:ovres}

We compare our MGCA-Net with state-of-the-art OV-TAL methods by calculating the average mAP across different tIoUs. Tab.~\ref{tab:ov_main} presents the performance comparison on THUMOS'14 and ActivityNet-1.3. Our MGCA-Net achieves the best results in terms of mAP$_{\text{base}}$, mAP$_{\text{novel}}$, and mAP$_{\text{all}}$ under both the 75\%-25\% and 50\%-50\% settings. Specifically, the performance improvement in mAP$_{\text{base}}$ demonstrates the advantages of the conventional classifier and action presence predictor compared with existing methods. For mAP$_{\text{novel}}$, benefiting from our proposed coarse-to-fine classifier, significant performance gains are attained under all settings. For instance, under all settings, it achieves over 6\% and 3\% performance improvements on THUMOS'14 and ActivityNet-1.3, respectively.

\begin{table*}[t!]
    \centering
    \caption{Performance comparison of ZS-TAL with state-of-the-art methods on THUMOS'14 and ActivityNet-1.3. "Avg." denotes the average across different tIoU thresholds. \textbf{Bold} values indicate the best results, and \underline{underlined} values indicate the second-best results.}
    \resizebox{1.0\linewidth}{!}{\begin{tabular}{clc|cc|cccccccccc}
\toprule
\multirow{2}{*}{\textbf{Split}} & \multicolumn{1}{c}{\multirow{2}{*}{\textbf{Method}}} & \multirow{2}{*}{\textbf{Venue}} & \multirow{2}{*}{\textbf{\begin{tabular}[c]{@{}c@{}}Prompt\\ Tuning\end{tabular}}} & \multirow{2}{*}{\textbf{\begin{tabular}[c]{@{}c@{}}Text\\ Feature\end{tabular}}} & \multicolumn{6}{c}{\textbf{THUMOS'14}} & \multicolumn{4}{c}{\textbf{ActivityNet-1.3}} \\ \cmidrule{6-15} 
 & \multicolumn{1}{c}{} &  &  &  & \textbf{0.3} & \textbf{0.4} & \textbf{0.5} & \textbf{0.6} & \textbf{0.7} & \multicolumn{1}{c|}{\textbf{Avg.}} & \textbf{0.5} & \textbf{0.75} & \textbf{0.95} & \textbf{Avg.} \\ \midrule
\multirow{12}{*}{\begin{tabular}[c]{@{}c@{}}75\% Seen\\ 25\% Unseen\end{tabular}} & B-II~\cite{stable2022eccv} & - & \ding{51} & CLIP-B & 28.5 & 20.3 & 17.1 & 10.5 & 6.9 & \multicolumn{1}{c|}{16.6} & 32.6 & 18.5 & 5.8 & 19.6 \\
 & B-I~\cite{stable2022eccv} & - & \ding{51} & CLIP-B & 33.0 & 25.5 & 18.3 & 11.6 & 5.7 & \multicolumn{1}{c|}{18.8} & 35.6 & 20.4 & 2.1 & 20.2 \\
 & Eff-Prompt~\cite{effprompt2022eccv} & ECCV'22 & \ding{51} & CLIP-B & 39.7 & 31.6 & 23.0 & 14.9 & 7.5 & \multicolumn{1}{c|}{23.3} & 37.6 & 22.9 & 3.8 & 23.1 \\
 & STABLE~\cite{stable2022eccv} & ECCV'22 & \ding{51} & CLIP-B & 40.5 & 32.3 & 23.5 & 15.3 & 7.6 & \multicolumn{1}{c|}{23.8} & 38.2 & 25.2 & 6.0 & 24.9 \\
 & UnLoc-B~\cite{unloc2023iccv} & ICCV'23 & \ding{51} & CLIP-B & - & - & - & - & - & \multicolumn{1}{c|}{-} & 36.9 & - & - & - \\
 & ZEETAD~\cite{zeetad2024wacv} & WACV'24 & \ding{51} & CLIP-B & 61.4 & 53.9 & 44.7 & 34.5 & 20.5 & \multicolumn{1}{c|}{43.2} & 51.0 & 33.4 & 5.9 & 32.5 \\
 & mProTEA~\cite{mprotea2024tcsvt} & TCSVT'24 & \ding{51} & CLIP-B & 43.1 & 38.2 & 28.2 & 18.1 & 8.7 & \multicolumn{1}{c|}{27.9} & 44.5 & 27.4 & \underline{7.9} & 27.6 \\
 & OVFormer~\cite{ovformer2024bmvc} & BMVC'24 & \ding{55} & DINOv2 & 49.8 & 43.8 & 35.8 & 27.8 & 19.2 & \multicolumn{1}{c|}{35.3} & 46.7 & 29.4 & 6.1 & 29.5 \\
 & DeTAL~\cite{detal2024tpami} & TPAMI'24 & \ding{51} & CLIP-B & 39.8 & 33.6 & 25.9 & 17.4 & 9.9 & \multicolumn{1}{c|}{25.3} & 39.3 & 26.4 & 5.0 & 25.8 \\
 & Ti-FAD~\cite{tifad2024neurips} & NeurIPS'24 & \ding{55} & CLIP-B & \underline{64.0} & \underline{58.5} & \textbf{49.7} & \underline{37.7} & \underline{24.1} & \multicolumn{1}{c|}{\underline{46.8}} & \textbf{53.8} & \underline{34.8} & 7.0 & \underline{34.7} \\
 & STOV-TAL~\cite{stovtal2025wacv} & WACV'25 & \ding{55} & CLIP-B & 47.8 & 39.1 & 28.4 & 17.6 & 9.1 & \multicolumn{1}{c|}{28.4} & 47.0 & 28.1 & 1.6 & 27.9 \\
 & MGCA-Net (Ours) & - & \ding{55} & CLIP-B & \textbf{66.4} & \textbf{59.7} & \underline{49.6} & \textbf{38.2} & \textbf{26.2} & \multicolumn{1}{c|}{\textbf{48.0}} & \underline{52.8} & \textbf{35.6} & \textbf{8.2} & \textbf{35.0} \\ \midrule
\multirow{12}{*}{\begin{tabular}[c]{@{}c@{}}50\% Seen\\ 50\% Unseen\end{tabular}} & B-II~\cite{stable2022eccv} & - & \ding{51} & CLIP-B & 21.0 & 16.4 & 11.2 & 6.3 & 3.2 & \multicolumn{1}{c|}{11.6} & 25.3 & 13.0 & 3.7 & 12.9 \\
 & B-I~\cite{stable2022eccv} & - & \ding{51} & CLIP-B & 27.2 & 21.3 & 15.3 & 9.7 & 4.8 & \multicolumn{1}{c|}{15.7} & 28.0 & 16.4 & 1.2 & 16.0 \\
 & Eff-Prompt~\cite{effprompt2022eccv} & ECCV'22 & \ding{51} & CLIP-B & 37.2 & 29.6 & 21.6 & 14.0 & 7.2 & \multicolumn{1}{c|}{21.9} & 32.0 & 19.3 & 2.9 & 19.6 \\
 & STABLE~\cite{stable2022eccv} & ECCV'22 & \ding{51} & CLIP-B & 38.3 & 30.7 & 21.2 & 13.8 & 7.0 & \multicolumn{1}{c|}{22.2} & 32.1 & 20.7 & 5.9 & 20.5 \\
 & UnLoc-B~\cite{unloc2023iccv} & ICCV'23 & \ding{51} & CLIP-B & - & - & - & - & - & \multicolumn{1}{c|}{-} & 40.2 & - & - & - \\
 & ZEETAD~\cite{zeetad2024wacv} & WACV'24 & \ding{51} & CLIP-B & 45.2 & 38.8 & 30.8 & 22.5 & 13.7 & \multicolumn{1}{c|}{30.2} & 39.2 & 25.7 & 3.1 & 24.9 \\
 & mProTEA~\cite{mprotea2024tcsvt} & TCSVT'24 & \ding{51} & CLIP-B & 41.2 & 36.3 & 26.3 & 16.8 & 8.4 & \multicolumn{1}{c|}{26.1} & 41.8 & 24.6 & \underline{6.1} & 25.6 \\
 & OVFormer~\cite{ovformer2024bmvc} & BMVC'24 & \ding{55} & DINOv2 & 42.8 & 37.3 & 30.6 & \underline{23.5} & 15.9 & \multicolumn{1}{c|}{30.5} & 42.8 & 27.3 & 6.0 & 27.2 \\
 & DeTAL~\cite{detal2024tpami} & TPAMI'24 & \ding{51} & CLIP-B & 38.3 & 32.3 & 24.4 & 16.3 & 9.0 & \multicolumn{1}{c|}{24.1} & 34.4 & 23.0 & 4.0 & 22.4 \\
 & Ti-FAD~\cite{tifad2024neurips} & NeurIPS'24 & \ding{55} & CLIP-B & \underline{57.0} & \underline{51.4} & \textbf{43.3} & \textbf{33.0} & \underline{21.2} & \multicolumn{1}{c|}{\underline{41.2}} & \textbf{50.6} & \underline{32.2} & 5.2 & \underline{32.0} \\
 & STOV-TAL~\cite{stovtal2025wacv} & WACV'25 & \ding{55} & CLIP-B & 44.2 & 35.7 & 25.7 & 16.5 & 8.0 & \multicolumn{1}{c|}{26.0} & 42.1 & 25.0 & 1.3 & 24.8 \\
 & MGCA-Net (Ours) & - & \ding{55} & CLIP-B & \textbf{58.0} & \textbf{51.6} & \underline{42.6} & \textbf{33.0} & \textbf{21.5} & \multicolumn{1}{c|}{\textbf{41.3}} & \underline{48.8} & \textbf{32.8} & \textbf{7.0} & \textbf{32.2} \\ \bottomrule
\end{tabular}}
\label{tab:zs_main}
\end{table*}

\subsection{Zero-Shot Results}\label{subsec:zsres}

Our MGCA-Net is not only applicable to the OV-TAL task but also to the ZS-TAL task when only novel action categories are considered. Specifically, for the ZS-TAL task, MGCA-Net removes the conventional classifier and action presence predictor, and only uses the coarse-to-fine classifier to predict novel action categories. Under this setup, the performance of MGCA-Net on novel actions can be measured more accurately.

In Tab.~\ref{tab:zs_main}, we present the performance comparison for the ZS-TAL task. Our MGCA-Net achieves the best average mAP under both the 75\%-25\% and 50\%-50\% settings on THUMOS'14 and ActivityNet-1.3. Notably, MGCA-Net exhibits excellent performance at high tIoU thresholds, which is crucial for accurate localization. For instance, under the 75\%-25\% setting, the average mAP of Ti-FAD is close to that of our MGCA-Net; however, compared to Ti-FAD, MGCA-Net improves by 2.1\% at the 0.7 tIoU threshold on THUMOS'14 and by 1.2\% at the 0.95 tIoU threshold on ActivityNet-1.3. This fully demonstrates the advantages of our proposed MGCA-Net. Additionally, like Ti-FAD and STOV-TAL, MGCA-Net does not require additional prompt tuning, ensuring that VLMs' original generalization ability is not compromised.

\begin{table}[t!]
    \centering
    \caption{Ablations on the conventional classifier and action presence predictor.}
    \resizebox{0.9\linewidth}{!}{\begin{tabular}{cl|ccc}
\toprule
\multicolumn{2}{l|}{\textbf{Conventional Classifier}} & \ding{55} & \ding{51} & \ding{51} \\
\multicolumn{2}{l|}{\textbf{Action Presence Predictor}} & \ding{55} & \ding{55} & \ding{51} \\ \midrule
\multicolumn{1}{c|}{\multirow{3}{*}{\begin{tabular}[c]{@{}c@{}}75\% Seen\\ 25\% Unseen\end{tabular}}} & mAP$_{base}$ & 52.0 & 57.9 & \textbf{59.2} \\
\multicolumn{1}{c|}{} & mAP$_{novel}$ & 33.9 & 33.7 & \textbf{34.0} \\
\multicolumn{1}{c|}{} & mAP$_{all}$ & 47.5 & 51.8 & \textbf{52.9} \\ \midrule
\multicolumn{1}{c|}{\multirow{3}{*}{\begin{tabular}[c]{@{}c@{}}50\% Seen\\ 50\% Unseen\end{tabular}}} & mAP$_{base}$ & 51.6 & 56.0 & \textbf{56.3} \\
\multicolumn{1}{c|}{} & mAP$_{novel}$ & 30.4 & 30.3 & \textbf{31.3} \\
\multicolumn{1}{c|}{} & mAP$_{all}$ & 41.0 & 43.2 & \textbf{43.9} \\ \bottomrule
\end{tabular}}
\label{tab:ablation_cc_app}
\end{table}

\subsection{Ablation Study}\label{subsec:ablationstudy}

In this subsection, we conduct a series of ablation studies to verify both the effectiveness of each module in MGCA-Net and the impact of different hyperparameters on localization performance.

\textbf{Ablations on the conventional classifier and action presence predictor.} As shown in Tab.~\ref{tab:ablation_cc_app}, we validate the effectiveness of the conventional classifier and action presence predictor modules on the OV-TAL task of the THUMOS'14 dataset. When neither module is used (first column of the performance comparison), we use VLMs to classify base and novel action categories simultaneously. It can be observed that in this case, the performance of mAP$_{\text{base}}$ decreases significantly. When only the conventional classifier is used (second column of the performance comparison), in Algo.~\ref{algo:generate_bnp}, we distinguish between base and novel proposals solely based on the predicted probabilities of base actions. It is evident that the introduction of the conventional classifier significantly improves the performance of mAP$_{\text{base}}$. However, base action probabilities tend to classify proposals as base actions, decreasing mAP$_{\text{novel}}$. When both modules are used (third column of the performance comparison), since the action presence predictor can better identify proposals where actions are likely to occur, both mAP$_{\text{base}}$ and mAP$_{\text{novel}}$ achieve significant improvements.

\begin{table}[t!]
    \centering
    \caption{Ablations on the coarse-to-fine classifier. For OV-TAL, we report mAP$_{\text{novel}}$; for ZS-TAL, we report the average mAP across tIoU thresholds [0.3:0.1:0.7].}
    \resizebox{1.0\linewidth}{!}{\begin{tabular}{c|l|cc}
\toprule
\multirow{2}{*}{\textbf{Split}} & \multicolumn{1}{c|}{\multirow{2}{*}{\textbf{Method}}} & \textbf{OV-TAL} & \textbf{ZS-TAL} \\ \cmidrule{3-4} 
 & \multicolumn{1}{c|}{} & \textbf{mAP$_{novel}$} & \textbf{Avg.} \\ \midrule
\multirow{3}{*}{\begin{tabular}[c]{@{}c@{}}75\% Seen\\ 25\% Unseen\end{tabular}} & coarse single template & 32.3 & 44.5 \\
 & coarse multi template fusion & 33.6 & 46.8 \\
 & coarse-to-fine classifier & \textbf{34.0} & \textbf{48.0} \\ \midrule
\multirow{3}{*}{\begin{tabular}[c]{@{}c@{}}50\% Seen\\ 50\% Unseen\end{tabular}} & coarse single template & 28.7 & 36.9 \\
 & coarse multi template fusion & 30.7 & 39.4 \\
 & coarse-to-fine classifier & \textbf{31.3} & \textbf{41.3} \\ \bottomrule
\end{tabular}}
\label{tab:ablataion_coarse2fine}
\end{table}

\begin{table}[t!]
    \centering
    \caption{Ablations on forms of ground truth for the action presence predictor, on the OV-TAL task of the THUMOS'14 dataset. \textit{fusion} denotes taking the average of the classification score and tIoU.}
    \resizebox{1.0\linewidth}{!}{\begin{tabular}{c|l|ccc}
\toprule
\textbf{Split} & \multicolumn{1}{c|}{\textbf{Type}} & \textbf{mAP$_{base}$} & \textbf{mAP$_{novel}$} & \textbf{mAP$_{all}$} \\ \midrule
\multirow{3}{*}{\begin{tabular}[c]{@{}c@{}}75\% Seen\\ 25\% Unseen\end{tabular}} & classification score & 58.0 & 33.0 & 51.8 \\
 & tIoU & \textbf{59.2} & \textbf{34.0} & \textbf{52.9} \\
 & fusion & 58.2 & 33.6 & 52.1 \\ \midrule
\multirow{3}{*}{\begin{tabular}[c]{@{}c@{}}50\% Seen\\ 50\% Unseen\end{tabular}} & classification score & 55.6 & 30.1 & 42.9 \\
 & tIoU & \textbf{56.3} & \textbf{31.3} & \textbf{43.9} \\
 & fusion & 55.8 & 31.1 & 43.5 \\ \bottomrule
\end{tabular}}
\label{tab:ablation_gt_app}
\end{table}

\textbf{Ablations on the coarse-to-fine classifier.} We analyze the effectiveness of the coarse-to-fine classifier on both the OV-TAL and ZS-TAL tasks using the THUMOS'14 dataset, with results presented in Tab.~\ref{tab:ablataion_coarse2fine}. In Tab.~\ref{tab:ablataion_coarse2fine}, \textit{coarse single template} denotes computing coarse categories based on a single template as described in Sec.~\ref{subsec:ctfc}, subsequently assigning all coarse categories to each novel proposal. This approach is widely used in the post-processing of most TAL methods~\cite{actionformer2022, tridet2023, bdrcnet2025} and some OV-TAL methods (e.g., STABLE~\cite{stable2022eccv}). \textit{Coarse multi template fusion} refers to obtaining coarse categories using the fused results of multiple templates. From Tab.~\ref{tab:ablataion_coarse2fine}, we can conclude that \textit{coarse multi template fusion} significantly improves detection performance. Additionally, our \textit{coarse-to-fine classifier} avoids using the aforementioned post-processing methods, while identifying the presence of actions at the video granularity and assigning action categories at the proposal granularity, further improving detection performance.

\textbf{Ablations on different ground truths of the action presence predictor.} In Sec.~\ref{subsec:app}, we use tIoU as the ground truth for the corresponding proposal in the action presence predictor. As shown in Tab.~\ref{tab:ablation_gt_app}, we compare other types of ground truth with tIoU on THUMOS'14. First, we use \textit{classification score}—i.e., the classification score of each proposal from the conventional classifier—as the ground truth. Since the \textit{classification score} is mainly trained on existing annotated data and fails to focus on novel proposals, there is a significant drop in localization performance. When using \textit{fusion} (i.e., fusing the average of the classification score and tIoU), the \textit{classification score} also exerts a negative impact on tIoU. In contrast, due to its category-agnostic nature, tIoU only focuses on localization quality and thus achieves the best performance.

\begin{table}[t!]
    \centering
    \caption{Ablations on $\lambda_{retain}$.}
    \resizebox{0.95\linewidth}{!}{\begin{tabular}{c|c|ccc}
\toprule
\textbf{Split} & $\bm{\lambda_{retain}}$ & \textbf{mAP$_{base}$} & \textbf{mAP$_{novel}$} & \textbf{mAP$_{all}$} \\ \midrule
\multirow{3}{*}{\begin{tabular}[c]{@{}c@{}}75\% Seen\\ 25\% Unseen\end{tabular}} & 0.4 & 58.7 & 33.7 & 52.5 \\
 & 0.5 & \textbf{59.2} & \textbf{34.0} & \textbf{52.9} \\
 & 0.6 & 58.5 & 33.2 & 52.2 \\ \midrule
\multirow{3}{*}{\begin{tabular}[c]{@{}c@{}}50\% Seen\\ 50\% Unseen\end{tabular}} & 0.4 & 56.0 & 31.1 & 43.5 \\
 & 0.5 & \textbf{56.3} & \textbf{31.3} & \textbf{43.9} \\
 & 0.6 & 55.6 & 30.5 & 43.1 \\ \bottomrule
\end{tabular}}
\label{tab:ablation_lambda_retain}
\end{table}

\begin{table}[t!]
    \centering
    \caption{Ablations on $\lambda_{base}$.}
    \resizebox{0.95\linewidth}{!}{\begin{tabular}{c|c|ccc}
\toprule
\textbf{Split} & $\bm{\lambda_{base}}$ & \textbf{mAP$_{base}$} & \textbf{mAP$_{novel}$} & \textbf{mAP$_{all}$} \\ \midrule
\multirow{3}{*}{\begin{tabular}[c]{@{}c@{}}75\% Seen\\ 25\% Unseen\end{tabular}} & 0.4 & \textbf{59.9} & 31.5 & 52.8 \\
 & 0.5 & 59.2 & 34.0 & \textbf{52.9} \\
 & 0.6 & 54.9 & \textbf{34.4} & 49.7 \\ \midrule
\multirow{3}{*}{\begin{tabular}[c]{@{}c@{}}50\% Seen\\ 50\% Unseen\end{tabular}} & 0.4 & \textbf{57.4} & 28.1 & 43.5 \\
 & 0.5 & 56.3 & 31.3 & \textbf{43.9} \\
 & 0.6 & 53.0 & \textbf{31.7} & 42.3 \\ \bottomrule
\end{tabular}}
\label{tab:ablation_lambda_base}
\end{table}

\textbf{Ablations on $\bm{\lambda_{retain}}$.} As shown in Tab.~\ref{tab:ablation_lambda_retain}, we validate different values of $\lambda_{\text{retain}}$ on the OV-TAL task of THUMOS'14. The optimal localization performance is achieved when $\lambda_{\text{retain}}=0.5$. Reducing $\lambda_{\text{retain}}$ increases false localizations in the results, degrading localization performance. Conversely, increasing $\lambda_{\text{retain}}$ leads to omitting some localizations, resulting in the worst performance among the three values.

\textbf{Ablations on $\bm{\lambda_{base}}$.} As shown in Tab.~\ref{tab:ablation_lambda_base}, we validate different values of $\lambda_{\text{base}}$ on the OV-TAL task of THUMOS'14. The $\lambda_{\text{base}}$ parameter primarily concerns which proposals are classified into base action categories. At lower values, such as $\lambda_{\text{base}}=0.4$, the localization results tend to lean toward base actions, leading to a significant drop in mAP$_{\text{novel}}$. Conversely, at higher values, such as $\lambda_{\text{base}}=0.6$, the localization results lean more toward novel actions, resulting in a significant drop in mAP$_{\text{base}}$. As shown in Tab.~\ref{tab:ablation_lambda_base}, when $\lambda_{\text{base}}=0.5$, it achieves a better balance between the two, with mAP$_{\text{all}}$ attaining the optimal performance.

\begin{table}[t!]
    \centering
    \caption{Ablations on $N_{coarse}$.}
    \resizebox{0.95\linewidth}{!}{\begin{tabular}{c|c|cccc}
\toprule
\textbf{Split} & $\bm{N_{coarse}}$ & \textbf{0.3} & \textbf{0.5} & \textbf{0.7} & \textbf{Avg.} \\ \midrule
\multirow{3}{*}{\begin{tabular}[c]{@{}c@{}}75\% Seen\\ 25\% Unseen\end{tabular}} & 1 & 64.3 & 48.6 & 26.1 & 46.9 \\
 & 2 & \textbf{66.4} & \textbf{49.6} & \textbf{26.2} & \textbf{48.0} \\
 & 3 & 64.3 & 48.1 & 25.6 & 46.5 \\ \midrule
\multirow{3}{*}{\begin{tabular}[c]{@{}c@{}}50\% Seen\\ 50\% Unseen\end{tabular}} & 1 & 56.0 & 41.4 & 21.1 & 40.0 \\
 & 2 & \textbf{58.0} & \textbf{42.6} & \textbf{21.5} & \textbf{41.3} \\
 & 3 & 57.6 & 42.3 & \textbf{21.5} & 40.9 \\ \bottomrule
\end{tabular}}
\label{tab:ablation_n_coarse}
\end{table}

\textbf{Ablations on $\bm{N_{coarse}}$.} $N_{\text{coarse}}$ is used to determine the number of coarse categories in Sec.~\ref{subsec:ctfc}, and it dictates the recall rate for novel action categories. When $N_{\text{coarse}}$ takes a larger value, it increases the recall rate but decreases accuracy. We conducted ablation studies on the impact of different values of $N_{\text{coarse}}$ on localization performance on the ZS-TAL task of THUMOS'14. As shown in Tab.~\ref{tab:ablation_n_coarse}, when $N_{\text{coarse}}=1$, it indicates that only the novel action category with the highest probability is selected for each video, which reduces the recall rate and thus degrades localization performance. When $N_{\text{coarse}}=3$, more noisy predictions are introduced, degrading localization performance. A good balance is achieved when $N_{\text{coarse}}=2$, attaining the optimal localization performance.

\begin{table}[t!]
    \centering
    \caption{Ablations on $N_{neg}$.}
    \resizebox{0.95\linewidth}{!}{\begin{tabular}{c|c|cccc}
\toprule
\textbf{Split} & $\bm{N_{neg}}$ & \textbf{0.3} & \textbf{0.5} & \textbf{0.7} & \textbf{Avg.} \\ \midrule
\multirow{3}{*}{\begin{tabular}[c]{@{}c@{}}75\% Seen\\ 25\% Unseen\end{tabular}} & 2 & 65.4 & 48.7 & \textbf{26.3} & 47.4 \\
 & 3 & \textbf{66.4} & \textbf{49.6} & 26.2 & \textbf{48.0} \\
 & 4 & 64.7 & 48.0 & 25.8 & 46.7 \\ \midrule
\multirow{3}{*}{\begin{tabular}[c]{@{}c@{}}50\% Seen\\ 50\% Unseen\end{tabular}} & 2 & 57.3 & 41.9 & \textbf{21.5} & 40.7 \\
 & 3 & \textbf{58.0} & \textbf{42.6} & \textbf{21.5} & \textbf{41.3} \\
 & 4 & 57.2 & 41.8 & 21.3 & 40.6 \\ \bottomrule
\end{tabular}}
\label{tab:ablation_n_neg}
\end{table}

\textbf{Ablations on $\bm{N_{neg}}$.} $N_{\text{neg}}$ is used in the coarse-to-fine classifier to control the proportion of negative samples in contrastive learning, and this parameter primarily affects novel action categories. Thus, we conducted ablation studies on different values of $N_{\text{neg}}$ on the ZS-TAL task of THUMOS'14. As shown in Tab.~\ref{tab:ablation_n_neg}, when $N_{\text{neg}}=3$ (with a positive-to-negative sample ratio of 1:3), the optimal localization performance is achieved. Decreasing or increasing the ratio of positive to negative samples leads to a certain degree of performance degradation.

\begin{figure*}[!t]
\centering
\subcaptionbox{Cricket Bowling\label{subfig:cricketbowling}}{
    \includegraphics[width=1.0\linewidth]{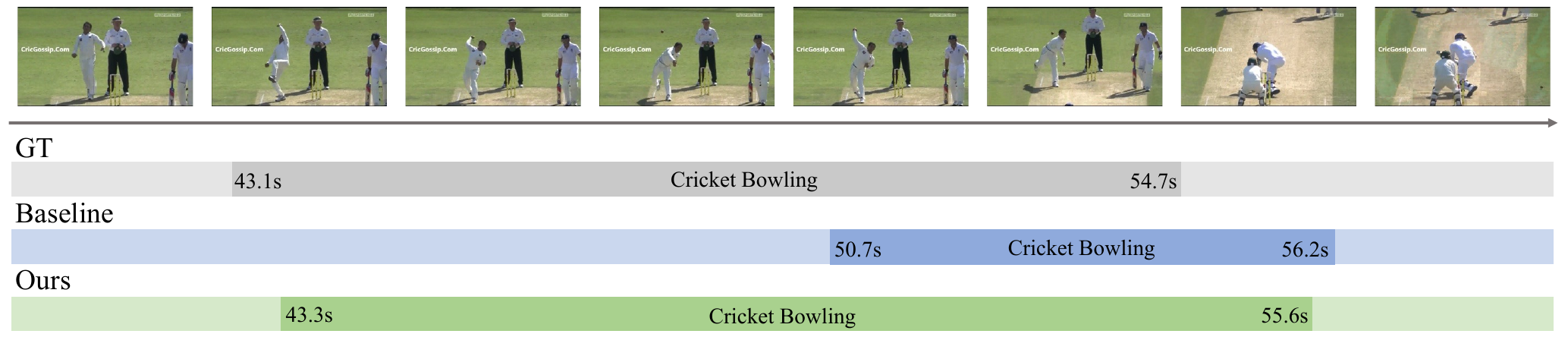}
}
\hfil
\subcaptionbox{GolfSwing\label{subfig:golfswing}}{
    \includegraphics[width=1.0\linewidth]{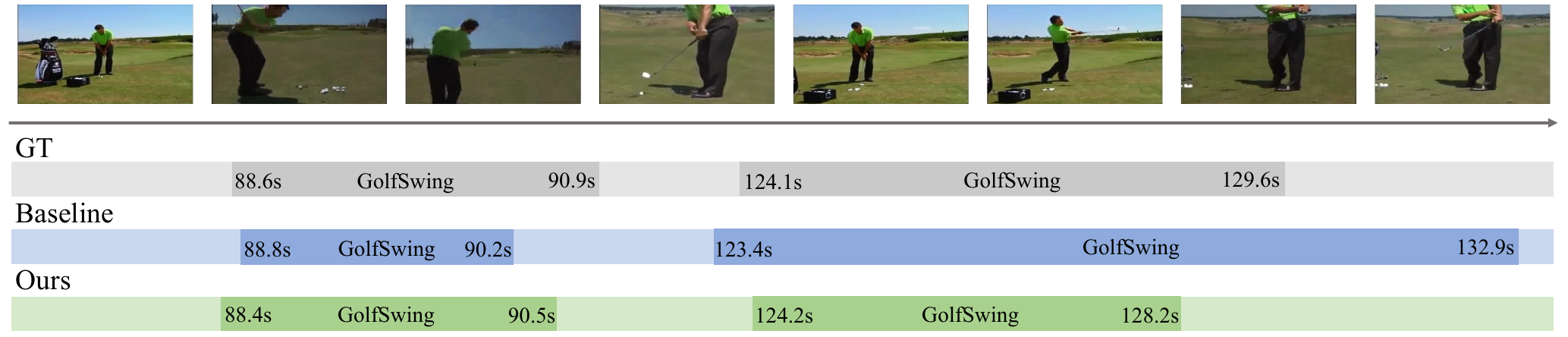}
}
\caption{Visualization of the localization results produced by the baseline and our MGCA-Net on THUMOS'14, where the ground truth labels are also provided. The baseline refers to our method with the conventional classifier, action presence predictor, and coarse-to-fine classifier removed—specifically, it uses a single template for action categories and predicts action categories based on the zero-shot capability of VLMs.}
\label{fig:viz}
\end{figure*}

\subsection{Qualitative Results}\label{subsec:qualitativeres}

To more intuitively demonstrate the performance of MGCA-Net, we visualize the localization results on the THUMOS'14 dataset. Specifically, Fig.~\ref{fig:viz} shows examples of localization results for the action categories \textit{Cricket Bowling} and \textit{Golf Swing}. In the comparison, we simultaneously compare MGCA-Net, the baseline, and the ground truth. Here, the baseline refers to our method with the conventional classifier, action presence predictor, and coarse-to-fine classifier removed. Specifically, it uses a single template for action categories and predicts action categories based on the zero-shot capability of VLMs. In Fig.~\ref{subfig:cricketbowling}, our method significantly outperforms the baseline. Furthermore, when multiple action instances exist in the input video, our proposed MGCA-Net also achieves superior localization performance, as shown in Fig.~\ref{subfig:golfswing}. Qualitative results across multiple scenarios further demonstrate the effectiveness of our proposed MGCA-Net.

\subsection{Limitations and Future work}\label{subsec:limitations}

Our proposed MGCA-Net employs the same category-agnostic localizer as existing methods~\cite{detal2024tpami, effprompt2022eccv} to predict category-agnostic action proposals. While this approach has achieved certain success, it overlooks the issue that unannotated action instances may exist in the training set. During the training of the localizer, these unannotated action instances are treated as negative samples, causing the localization results to be biased toward annotated action instances. This limits the generalization ability of the localizer to a certain extent, thereby exerting a negative impact on MGCA-Net.

In the future, we will explore generating pseudo-labels for unannotated action instances in the training set, so as to alleviate the aforementioned issue. The introduction of high-quality pseudo-labels enables the localizer to focus on more generalizable action instances, thereby enhancing its generalization ability.

\section{Conclusion}

In this paper, we propose the Multi-Grained Category-Aware Network (MGCA-Net) for open-vocabulary temporal action localization. MGCA-Net perceives action categories at multiple granularities, effectively alleviating the limitations of existing methods that rely on single-granularity action category perception. Specifically, MGCA-Net consists of a localizer, a conventional classifier, an action presence predictor, and a coarse-to-fine classifier. Herein, the localizer is used to localize category-agnostic action proposals. The conventional classifier and action presence predictor predict each action proposal's probabilities over all base categories and its likelihood of being an action instance. Using these two probabilities, they derive base action instances and novel proposals. Subsequently, the coarse-to-fine classifier identifies all action categories present in the video at the video granularity and assigns action categories to each novel proposal at the proposal granularity, thereby yielding novel action instances. The final localization results are the union of base and novel action instances. Extensive experiments on both OV-TAL and ZS-TAL tasks demonstrate the effectiveness of MGCA-Net.

\bibliographystyle{IEEEtran}
\bibliography{mgcanet}

\vfill

\end{document}